\documentclass[10pt]{article}

\usepackage[a4paper,margin=1in]{geometry}
\usepackage[utf8]{inputenc}
\usepackage[T1]{fontenc}
\usepackage{amsmath,amssymb,amsfonts}
\usepackage{booktabs}
\usepackage{graphicx}
\usepackage{hyperref}
\usepackage{microtype}
\usepackage{xcolor}
\usepackage[margin=1in]{geometry}
\usepackage{natbib}
\usepackage{caption}
\usepackage{float}
\usepackage{mathpazo}

\hypersetup{
    colorlinks=true,
    linkcolor={blue!70!black},
    citecolor={green!50!black},
    urlcolor={blue!70!black}
}

\title{Superposition Is Not Necessary: \\
\large A Mechanistic Interpretability Analysis of Transformer Representations \\ for Time Series Forecasting}

\author{
    Alper Y{\i}ld{\i}r{\i}m\thanks{Corresponding author: \texttt{yildirim.alper.dev@gmail.com} \\ ORCID: 0009-0004-8029-1772}
}

\date{}

\begin{document}

\maketitle

\begin{abstract}
Transformer architectures have been widely adopted for time series forecasting, yet whether the representational mechanisms that make them powerful in NLP actually engage on time series data remains unexplored. The persistent competitiveness of simple linear models such as DLinear has fueled ongoing debate, but no mechanistic explanation for this phenomenon has been offered. We address this gap by applying sparse autoencoders (SAEs), a tool from mechanistic interpretability, to probe the internal representations of PatchTST. We first establish that a single-layer, narrow-dimensional transformer matches the forecasting performance of deeper configurations across commonly used benchmarks. We then train SAEs on the post-GELU intermediate FFN activations with dictionary sizes ranging from $0.5\times$ to $4.0\times$ the native dimensionality. Expanding the dictionary yields negligible downstream performance change (average $0.214\%$), with large portions of overcomplete dictionaries remaining inactive. Targeted causal interventions on dominant latent features produce minimal forecast perturbation. Across all evaluated settings, we observe no empirical evidence that the analyzed FFN representations rely on strong superposition. Instead, the representations remain sparse, stable under aggressive dictionary expansion, and largely insensitive to latent interventions. These results demonstrate that superposition is not necessary for competitive performance on standard forecasting benchmarks, suggesting they may not demand the rich compositional representations that drive transformer success in language modeling, and helping explain the persistent competitiveness of simple linear models.
\end{abstract}

\vspace{0.5em}

% ============================================================
% 1. INTRODUCTION
% ============================================================
\section{Introduction}

The success of transformers in natural language processing \citep{vaswani2017attention} has motivated their widespread adoption for time series forecasting. Models such as Informer \citep{zhou2021informer}, Autoformer \citep{wu2021autoformer}, FEDformer \citep{zhou2022fedformer}, PatchTST \citep{nie2023patchtst}, and iTransformer \citep{liu2024itransformer} have progressively imported architectural innovations from the language modeling domain, each proposing new attention mechanisms or input representations tailored to temporal data. Yet a fundamental question remains largely unexamined: do the representational mechanisms that make transformers powerful in NLP --- particularly the ability to compose and compress rich features through superposition --- actually engage on time series forecasting tasks?

This question is not merely theoretical. \citet{zeng2023dlinear} demonstrated that DLinear, a simple single-layer linear model, matches or outperforms many transformer-based forecasters on standard benchmarks, igniting an ongoing debate about whether transformer complexity is justified for time series. Subsequent work has responded with improved architectures \citep{wu2023timesnet, liu2024itransformer}, yet the debate has remained confined to benchmark comparisons: each side proposes a model, reports metrics, and draws architectural conclusions. What has been conspicuously absent is \emph{mechanistic} evidence --- an examination of what transformers actually learn internally when applied to time series data, and whether this differs fundamentally from the representations they build for language.

In parallel, the field of mechanistic interpretability has developed powerful tools for understanding the internal representations of neural networks. The superposition hypothesis \citep{elhage2022superposition} proposes that neural networks represent more features than they have dimensions by encoding them in overlapping, superimposed directions. Sparse autoencoders (SAEs) have emerged as a primary tool for decomposing these superimposed representations into interpretable sparse features, with notable success in large language models \citep{bricken2023monosemanticity, templeton2024scaling, cunningham2023sparse}. In LLMs, SAE analysis consistently reveals dense superposition: expanding the dictionary uncovers progressively more monosemantic features, and intervening on individual latents produces measurable behavioral changes. However, these tools have seen virtually no application to time series transformers, leaving a significant gap in our understanding of how these models represent temporal patterns.

We bridge these two open questions by applying SAE-based mechanistic analysis to PatchTST \citep{nie2023patchtst}, the canonical channel-independent patch-based transformer for time series forecasting and a central reference point in the DLinear debate. Our choice of PatchTST is deliberate: it is not merely one architecture among many, but the model whose design most directly inherits the representational machinery of NLP transformers. If its internal representations do not rely on superposition to achieve competitive forecasting performance, this constitutes evidence that the task itself does not demand it, rather than reflecting a limitation of one particular architecture.

Our analysis proceeds in three steps. First, we establish that a single-layer, narrow-dimensional PatchTST --- modernized with rotary position embeddings \citep{su2021rope} and RMSNorm \citep{zhang2023rmsnorm} --- matches the forecasting performance of deeper configurations across eight commonly used benchmarks, demonstrating that minimal transformer capacity suffices for these tasks. Second, we train SAEs on the post-GELU intermediate FFN activations of these models with dictionary sizes ranging from $0.5\times$ to $4.0\times$ the native dimensionality. We find that expanding the dictionary yields negligible changes in downstream forecasting performance, that expanded dictionary capacity provides no functional benefit, and that targeted causal interventions on dominant latent features produce minimal forecast perturbation. Third, we synthesize these findings into a mechanistic argument: superposition is not necessary for competitive performance on standard time series forecasting benchmarks.

Crucially, our claim is not that no time series transformer could ever employ superposition, nor that superposition cannot exist within these models. Rather, we demonstrate that competitive forecasting performance is achievable without it, which suggests that the standard benchmarks driving the DLinear debate may not contain the representational richness that would require --- or reward --- the complex feature composition that transformers uniquely enable in language. This offers a mechanistic explanation, rather than a purely empirical one, for why simple linear models remain competitive: the underlying data may simply not demand what transformers are best at.

% ============================================================
% 2. RELATED WORK
% ============================================================
\section{Related Work}

\subsection{Transformers for Time Series Forecasting}

The application of transformer architectures to time series forecasting has progressed rapidly, beginning with Informer \citep{zhou2021informer}, which introduced sparse attention for long-sequence efficiency, followed by Autoformer \citep{wu2021autoformer} and FEDformer \citep{zhou2022fedformer}, which incorporated decomposition and frequency-domain mechanisms respectively. PatchTST \citep{nie2023patchtst} introduced channel-independent patching, treating each variate as a separate token sequence and segmenting it into subseries-level patches, achieving strong performance with a simpler design. More recently, iTransformer \citep{liu2024itransformer} inverted the standard approach by embedding across variates rather than time steps. Throughout this progression, each model has imported architectural innovations from NLP --- attention variants, positional encodings, normalization schemes --- without examining whether the representational mechanisms that make these components powerful in language actually engage on temporal data.

\subsection{The Linear Model Debate}

\citet{zeng2023dlinear} challenged the transformer paradigm by demonstrating that DLinear, a single-layer linear model with seasonal-trend decomposition, matches or outperforms transformer-based forecasters across standard benchmarks. This finding ignited an ongoing debate about whether transformer complexity is justified for time series. Subsequent work has responded with improved architectures \citep{wu2023timesnet, liu2024itransformer}, and practitioners have noted that transformers offer advantages when covariates and cross-variate dependencies are important.
More recently, \citet{xu2024fits} pushed this finding considerably further:
their FITS model matches PatchTST across standard long-term forecasting
benchmarks using only ${\sim}10$k--$50$k parameters via a single
complex-valued linear layer in the frequency domain --- more than two
orders of magnitude smaller than transformer baselines, and roughly an
order of magnitude smaller than DLinear itself.
 However, the debate has remained confined to benchmark comparisons: each side proposes a model, reports metrics, and draws architectural conclusions. No prior work has offered \emph{mechanistic} evidence --- an examination of what transformers actually learn internally --- to explain why simple linear models remain competitive on standard forecasting benchmarks.

\subsection{Mechanistic Interpretability and Sparse Autoencoders}

The superposition hypothesis \citep{elhage2022superposition} proposes that neural networks represent more features than they have dimensions by encoding them in overlapping, superimposed directions. Sparse autoencoders (SAEs) have emerged as the primary tool for decomposing these superimposed representations into interpretable sparse features. In large language models, SAE analysis has consistently revealed dense superposition: expanding the dictionary uncovers progressively more monosemantic features, and intervening on individual latents produces measurable behavioral changes \citep{bricken2023monosemanticity, templeton2024scaling, cunningham2023sparse}. SAEs have also been applied to vision transformers, where they successfully extract interpretable visual features \citep{pach2025vision_sae}, establishing that the methodology transfers beyond language.

Work on understanding the internal structure of NLP transformers extends beyond superposition analysis. \citet{tenney2019bert} demonstrated that BERT's layers resolve linguistic abstractions in a progression mirroring the classical NLP pipeline --- from POS tagging in early layers to coreference in later layers --- providing evidence that depth serves a clear functional purpose in language models. \citet{niu2022does} subsequently challenged the simplicity of this pipeline interpretation, showing through their GridLoc probe that BERT's linguistic structure is distributed across both layers and token positions in a more complex pattern than a strict pipeline would predict. In either interpretation, NLP transformers exhibit rich internal representational structure that requires and utilizes multiple layers.

An important methodological consideration is raised by \citet{heap2025sae_random}, who demonstrated that SAEs can extract seemingly interpretable features even from randomly initialized, untrained transformers. This finding suggests that some SAE results may reflect statistical properties of input data rather than learned computations. Our work addresses this concern by focusing on \emph{downstream functional impact} --- dictionary scaling effects on forecasting performance and causal intervention effects on model outputs --- rather than relying solely on feature interpretability.

The most closely related concurrent work applies SAEs to large pretrained time series foundation models: \citet{mishra2026chronos} on Chronos-T5-Large and \citet{divo2026zeitgeist} on Chronos-2, both recovering interpretable feature structure within these models. \citet{kalnare2026mechanistic} apply attribution-based analysis to transformer-based time series \emph{classification} without using SAEs. Our work targets a complementary question: whether superposition is \emph{necessary} for competitive forecasting performance on the standard benchmarks driving the DLinear debate. We study a small, task-specific transformer rather than a pretrained foundation model, and probe for superposition through dictionary scaling and causal interventions rather than characterizing learned features --- finding, in contrast to TSFM work, no superposition signatures.
% ============================================================
% 3. METHODOLOGY
% ============================================================
\section{Methodology}

\subsection{Base Architecture}

We adopt PatchTST \citep{nie2023patchtst} as our base forecasting model, retaining its core channel-independent patching design: each input variate is treated as an independent univariate sequence, segmented into overlapping patches of length 16 with stride 8, and projected into a shared transformer backbone. The model applies Reversible Instance Normalization (RevIN) \citep{kim2022revin} at the input and output boundaries to handle distribution shift.

We make two architectural modernizations relative to the original PatchTST configuration. First, we replace the learned absolute positional embeddings with Rotary Position Embeddings (RoPE) \citep{su2021rope}, which encode relative position directly within the attention computation. This choice is motivated both by improved position generalization and by compatibility with SAE analysis: RoPE avoids entangling positional information into the residual stream, producing cleaner activation distributions at the FFN layer. Second, we replace BatchNorm with RMSNorm \citep{zhang2023rmsnorm}, following modern transformer practice.

Critically, we use a single transformer layer (\texttt{depth=1}) with narrow model dimensions. The per-dataset $d_\text{model}$ values range from 16 (Exchange, ETT variants) to 128 (Electricity), with an FFN intermediate dimension of $2 \times d_\text{model}$ via a standard GELU-activated feed-forward network. The original PatchTST paper evaluated depths of 3 to 6 layers but did not explore minimal configurations. As we demonstrate in Section~\ref{sec:single_layer}, this single-layer configuration matches published PatchTST performance across all eight benchmarks, establishing that minimal transformer capacity suffices for these tasks.

\subsection{Training Protocol}

We evaluate on eight commonly used long-term forecasting benchmarks: Weather, Electricity, and Traffic~\citep{wu2021autoformer}, Exchange~\citep{lai2018modeling}, and ETTh1/h2/m1/m2~\citep{zhou2021informer}, obtained via the Time Series Library. For the ETT datasets, we use the standard fixed splits (train/val/test). For all other datasets, we apply a 70/10/20 chronological split. All variates are normalized using a StandardScaler fitted on the training partition only.

Models are trained with a lookback window of 336 time steps and evaluated at four prediction horizons: 96, 192, 336, and 720. We use AdamW with a learning rate of $2 \times 10^{-4}$ and no weight decay, MSE loss, dropout of 0.2, and gradient clipping at norm 1.0. Training proceeds for up to 80 epochs with a patience of 15 on validation loss, using ReduceLROnPlateau scheduling (factor 0.5, patience 3). All training is performed with bfloat16 mixed precision. Each configuration is trained with a fixed random seed.

\subsection{Sparse Autoencoder Design}

To probe for superposition within the transformer's feed-forward network, we train sparse autoencoders (SAEs) on the post-GELU intermediate FFN activations. The hook is placed at the output of the GELU nonlinearity, capturing the intermediate representation after the up-projection and activation but before the down-projection. 
This is the standard hook location used in LLM SAE 
studies \citep{bricken2023monosemanticity, 
cunningham2023sparse}. We verify that these activations 
are functionally active --- rather than an inert component 
--- via zero ablation in Appendix~\ref{app:ffn_ablation}.
The SAE architecture consists of a linear encoder mapping from the FFN intermediate dimension $d_\text{ff}$ to a latent dimension $d_\text{hidden}$, followed by a ReLU activation, and a linear decoder mapping back to $d_\text{ff}$. The decoder columns are constrained to unit norm after each optimization step, following standard practice \citep{bricken2023monosemanticity}. We train SAEs at three dictionary scales: $0.5\times$, $1.0\times$, and $4.0\times$ the native FFN dimensionality, producing undercomplete, matched, and overcomplete dictionaries respectively.

Activations are harvested by running the frozen base model over training data, collecting post-GELU outputs up to a maximum of $10^6$ patch-level activation vectors. The SAE is then trained on these harvested activations using Adam with a learning rate of $10^{-3}$, minimizing a combined reconstruction and sparsity objective:
\begin{equation}
    \mathcal{L} = \| \mathbf{x} - \hat{\mathbf{x}} \|_2^2 + \lambda \| \mathbf{f} \|_1
\end{equation}
where $\mathbf{x}$ is the original activation, $\hat{\mathbf{x}}$ is the SAE reconstruction, $\mathbf{f}$ is the latent activation vector, and $\lambda = 0.01$. Training runs for up to 50 epochs with early stopping (patience of 3, improvement threshold of 0.1\%).

\subsection{Evaluation Protocol}
\label{sec:eval_protocol}

Our evaluation targets three distinct signatures of superposition, each designed to test a specific prediction that strong sparse-feature compression would produce:

\paragraph{Dictionary Scaling Analysis.} If the FFN activations contain substantial superposed structure, expanding the SAE dictionary should progressively disentangle hidden features, improving reconstruction fidelity and preserving downstream behavior. We test this by replacing the original post-GELU activations with SAE reconstructions during inference and measuring the resulting forecasting degradation (percentage change in test MSE) across all dictionary scales. Under strong superposition, we would expect the $4.0\times$ dictionary to substantially outperform the $0.5\times$ dictionary. A flat degradation profile across scales indicates that additional dictionary capacity is not uncovering meaningful hidden structure.

\paragraph{Dead Latent Analysis.} If the model is compressing many sparse features into limited dimensions, an overcomplete dictionary should activate a large number of additional latents to represent the newly disentangled features. We measure dead latent rates in the $4.0\times$ SAE by tracking which latents exceed an activation threshold of $10^{-5}$ across the full test set. High dead latent rates indicate that the expanded dictionary capacity is not being utilized, suggesting that the native representations do not contain large quantities of compressed sparse features awaiting recovery.

\paragraph{Causal Latent Interventions.} If the SAE has recovered functionally important sparse features, intervening on individual latents should produce measurable downstream effects. For each benchmark, we identify the top 10 most highly activated latents in the $4.0\times$ dictionary (ranked by total activation magnitude across the full test set) and amplify each by a factor of $5.0\times$ during inference. The resulting prediction shift is measured as the MAE between the original and intervened forecasts. Small intervention effects suggest that individual sparse features do not serve as dominant causal directions controlling the forecast output.

% ============================================================
% 4. RESULTS
% ============================================================
\section{Results}
\subsection{Single-Layer Sufficiency}
\label{sec:single_layer}

Before probing for superposition, we first establish that a shallow transformer with a unified hyperparameter configuration remains competitive on standard long-term forecasting benchmarks. Despite using only a single transformer layer, a shared hyperparameter configuration, and substantially narrower model dimensions, our model achieves broadly comparable performance to the published PatchTST/42 results \citep{nie2023patchtst}. While this minimal configuration serves primarily as a simplified model for our interpretability analysis, its strong performance indicates that it is a sufficient representative of the architecture's capabilities (see Appendix~\ref{app:single_layer} for full forecasting results). 

This observation motivates the central question of this work: if relatively shallow transformers already capture most of the achievable performance, what representational mechanisms --- if any --- does the FFN layer actually employ?

% --------------------------------------------------
\subsection{Minimal Sparse-Feature Scaling Under Dictionary Expansion}

This section reports results for the dictionary scaling and dead latent analyses described in Section~\ref{sec:eval_protocol}.

To test whether PatchTST FFN representations contain recoverable sparse structure hidden through superposition, we trained Sparse Autoencoders (SAEs) on post-GELU FFN activations with dictionary sizes ranging from $0.5\times$ to $4.0\times$ the native FFN dimensionality. During inference, the original activations were replaced with SAE reconstructions and downstream forecasting degradation was measured across all benchmarks and prediction horizons.

Despite the substantial increase in representational capacity, dictionary expansion produced only marginal downstream effects. Across all datasets and horizons, the average performance difference between the $0.5\times$ and $4.0\times$ dictionaries was only $0.214\%$. In many cases, degradation remained nearly constant across scales, with several benchmarks even showing slight improvements consistent with mild denoising effects.

Latent utilization statistics support this result. Dead latent rates varied across datasets, exceeding $60\%$ on Traffic while remaining below $7\%$ on most benchmarks. Together, these findings indicate that expanding sparse latent capacity does not uncover substantial additional functionally relevant structure within the analyzed FFN activations.

Unlike the strong sparse-feature scaling commonly reported in LLM SAE studies, the PatchTST representations examined here remain stable under aggressive dictionary expansion, suggesting weak sparse-feature pressure within the probed FFN layer. Although the $4.0\times$ dictionary consistently achieves lower reconstruction MSE than the $0.5\times$ dictionary (Table~\ref{tab:sae_fidelity}), this improved internal fidelity does not translate into downstream forecasting gains.

A potential concern is that most datasets still show $95\%+$ of $4.0\times$ latents firing above threshold. However, activity alone is not diagnostic: high latent activity combined with negligible downstream scaling ($0.214\%$) and minimal causal sensitivity ($\approx 0.031$ MAE; Section~\ref{sec:causal_interventions}) suggests these latents capture reconstruction-relevant but functionally inert structure rather than superposed features awaiting disentanglement.

% --------------------------------------------------

\begin{table}[H]
\centering
\caption{Sparse autoencoder analysis across eight standard forecasting benchmarks. Performance degradation remains stable despite $800\%$ expansion of the latent dictionary, and causally amplifying the top-10 most active latents by $500\%$ yields negligible downstream shift, suggesting that superposition is not necessary for competitive forecasting performance. Full SAE fidelity metrics (L0 sparsity and reconstruction MSE) are reported in Table~\ref{tab:sae_fidelity}.}
\resizebox{\textwidth}{!}{
\begin{tabular}{l c c c c c c c}
\toprule
\textbf{Dataset} & \textbf{Horizon} & \textbf{Base MSE} & \textbf{Dead Latents} & \textbf{0.5x Deg.} & \textbf{1.0x Deg.} & \textbf{4.0x Deg.} & \textbf{Causal Shift MAE} \\
\midrule
Weather & 96 & 0.1480 & 0.2\% & +0.42\% & +0.57\% & +0.23\% & 0.0293 \\
 & 192 & 0.1927 & 0.8\% & +0.02\% & $-$0.09\% & +0.07\% & 0.0292 \\
 & 336 & 0.2482 & 1.6\% & $-$0.03\% & $-$0.08\% & $-$0.08\% & 0.0208 \\
 & 720 & 0.3221 & 0.2\% & +0.00\% & $-$0.11\% & $-$0.16\% & 0.0297 \\
\midrule
Electricity & 96 & 0.1303 & 4.1\% & +3.28\% & +3.01\% & +2.89\% & 0.0446 \\
 & 192 & 0.1491 & 5.3\% & +2.36\% & +2.30\% & +2.11\% & 0.0508 \\
 & 336 & 0.1680 & 2.2\% & +2.82\% & +2.60\% & +2.59\% & 0.0614 \\
 & 720 & 0.2049 & 6.2\% & +1.99\% & +1.74\% & +1.71\% & 0.0561 \\
\midrule
Traffic & 96 & 0.3895 & 0.3\% & +0.49\% & +0.31\% & +0.31\% & 0.0517 \\
 & 192 & 0.4174 & 54.9\% & +1.79\% & +1.67\% & +1.32\% & 0.0542 \\
 & 336 & 0.4282 & 65.8\% & +2.99\% & +1.69\% & +1.23\% & 0.0516 \\
 & 720 & 0.4566 & 63.5\% & +1.78\% & +1.42\% & +1.10\% & 0.0395 \\
\midrule
Exchange & 96 & 0.0963 & 7.0\% & +0.98\% & +0.61\% & +0.93\% & 0.0127 \\
 & 192 & 0.1911 & 3.9\% & +0.28\% & +0.49\% & +0.63\% & 0.0375 \\
 & 336 & 0.3723 & 6.2\% & $-$0.49\% & $-$0.35\% & $-$0.20\% & 0.0183 \\
 & 720 & 0.9783 & 4.7\% & +1.71\% & +1.50\% & +1.38\% & 0.0336 \\
\midrule
ETTh1 & 96 & 0.3779 & 4.7\% & +0.01\% & $-$0.01\% & $-$0.00\% & 0.0218 \\
 & 192 & 0.4230 & 5.5\% & $-$0.09\% & $-$0.13\% & $-$0.11\% & 0.0223 \\
 & 336 & 0.4524 & 3.9\% & +0.04\% & $-$0.03\% & $-$0.02\% & 0.0251 \\
 & 720 & 0.4774 & 4.7\% & $-$0.28\% & $-$0.26\% & $-$0.24\% & 0.0254 \\
\midrule
ETTh2 & 96 & 0.2923 & 3.9\% & $-$0.48\% & $-$0.74\% & $-$0.45\% & 0.0254 \\
 & 192 & 0.3562 & 3.1\% & $-$0.47\% & $-$0.34\% & $-$0.44\% & 0.0240 \\
 & 336 & 0.3839 & 3.1\% & $-$0.94\% & $-$0.96\% & $-$0.93\% & 0.0312 \\
 & 720 & 0.4090 & 4.7\% & +0.04\% & $-$0.06\% & $-$0.02\% & 0.0175 \\
\midrule
ETTm1 & 96 & 0.3070 & 1.6\% & +0.37\% & +0.34\% & +0.28\% & 0.0223 \\
 & 192 & 0.3384 & 0.8\% & +0.43\% & +0.29\% & +0.37\% & 0.0234 \\
 & 336 & 0.3655 & 0.0\% & +0.14\% & +0.37\% & +0.39\% & 0.0205 \\
 & 720 & 0.4288 & 1.6\% & $-$0.21\% & $-$0.19\% & $-$0.15\% & 0.0216 \\
\midrule
ETTm2 & 96 & 0.1710 & 0.8\% & +0.01\% & +0.03\% & $-$0.06\% & 0.0213 \\
 & 192 & 0.2269 & 0.0\% & $-$0.33\% & $-$0.29\% & $-$0.36\% & 0.0230 \\
 & 336 & 0.2885 & 0.0\% & $-$0.02\% & $-$0.11\% & $-$0.07\% & 0.0175 \\
 & 720 & 0.3706 & 0.8\% & $-$1.00\% & $-$0.62\% & $-$0.74\% & 0.0240 \\
\bottomrule
\end{tabular}
}
\label{tab:sae_results}
\end{table}

\subsection{Limited Downstream Sensitivity to Latent Interventions}
\label{sec:causal_interventions}

This section reports results for the causal latent intervention protocol described in Section~\ref{sec:eval_protocol}.

To evaluate the functional influence of SAE-derived sparse features, we performed targeted causal interventions on the learned latent representations. For each dataset, we identified the top 10 most highly activated latents within the $4.0\times$ overcomplete SAE dictionary and amplified each latent by a factor of $5.0\times$ during inference on the full test set.

Despite these aggressive interventions, downstream forecasting behavior remained largely stable. Across all benchmarks, the average intervention-induced prediction shift remained approximately $0.031$ MAE, while the maximum observed shift reached only $0.061$. These comparatively small output changes persisted across all datasets and prediction horizons, despite the large magnitude of the latent perturbations.

The intervention results suggest that the dominant SAE features identified within the analyzed post-GELU FFN activations do not individually exert strong control over the final forecast outputs. Instead, the forecasting computation appears comparatively robust to localized sparse latent perturbations, potentially indicating distributed, redundancy-tolerant, or smoothly shared internal representations rather than highly dominant sparse causal directions. Although we cannot rule out that alternative SAE training regimes, sparsity penalties, or decomposition methods might recover stronger sparse structure, the present results provide no evidence for strong superposition within the probed representations.

% ============================================================
% MAIN RESULTS TABLE
% ============================================================

\begin{table}[H]
\centering
\caption{Internal sparse autoencoder fidelity metrics. The $4.0\times$ overcomplete dictionary consistently achieves lower reconstruction MSE than the $0.5\times$ dictionary, yet this improved internal fidelity does not translate into downstream forecasting gains (see Table~\ref{tab:sae_results}), suggesting the additionally recovered structure is not functionally relevant.}
\resizebox{\textwidth}{!}{
\begin{tabular}{l c c c c c c c}
\toprule
\textbf{Dataset} & \textbf{Horizon} & \textbf{L0 ($0.5\times$)} & \textbf{L0 ($1.0\times$)} & \textbf{L0 ($4.0\times$)} & \textbf{Recon MSE ($0.5\times$)} & \textbf{Recon MSE ($1.0\times$)} & \textbf{Recon MSE ($4.0\times$)} \\
\midrule
Weather & 96 & 4.1 & 4.8 & 6.3 & 0.0375 & 0.0339 & 0.0278 \\
 & 192 & 4.0 & 5.2 & 6.1 & 0.0334 & 0.0287 & 0.0262 \\
 & 336 & 2.2 & 2.7 & 3.1 & 0.0191 & 0.0174 & 0.0162 \\
 & 720 & 3.7 & 4.4 & 5.2 & 0.0265 & 0.0234 & 0.0228 \\
\midrule
Electricity & 96 & 2.5 & 2.9 & 3.2 & 0.0449 & 0.0428 & 0.0402 \\
 & 192 & 2.1 & 2.3 & 2.7 & 0.0355 & 0.0344 & 0.0320 \\
 & 336 & 1.8 & 1.9 & 2.1 & 0.0265 & 0.0259 & 0.0252 \\
 & 720 & 0.6 & 0.9 & 1.0 & 0.0207 & 0.0194 & 0.0190 \\
\midrule
Traffic & 96 & 1.5 & 2.1 & 2.7 & 0.0346 & 0.0289 & 0.0272 \\
 & 192 & 1.0 & 1.2 & 1.6 & 0.0278 & 0.0252 & 0.0213 \\
 & 336 & 0.3 & 1.2 & 1.6 & 0.0392 & 0.0239 & 0.0204 \\
 & 720 & 0.6 & 1.0 & 1.5 & 0.0301 & 0.0243 & 0.0198 \\
\midrule
Exchange & 96 & 3.5 & 4.7 & 6.8 & 0.0065 & 0.0054 & 0.0042 \\
 & 192 & 3.6 & 4.7 & 7.1 & 0.0071 & 0.0061 & 0.0056 \\
 & 336 & 3.6 & 4.1 & 7.3 & 0.0085 & 0.0068 & 0.0055 \\
 & 720 & 4.0 & 5.8 & 9.3 & 0.0113 & 0.0083 & 0.0066 \\
\midrule
ETTh1 & 96 & 3.4 & 4.7 & 7.0 & 0.0111 & 0.0087 & 0.0073 \\
 & 192 & 4.0 & 4.8 & 7.8 & 0.0113 & 0.0092 & 0.0071 \\
 & 336 & 4.5 & 5.8 & 9.5 & 0.0128 & 0.0097 & 0.0075 \\
 & 720 & 3.6 & 4.9 & 6.7 & 0.0101 & 0.0079 & 0.0065 \\
\midrule
ETTh2 & 96 & 4.6 & 6.2 & 10.4 & 0.0165 & 0.0117 & 0.0085 \\
 & 192 & 4.3 & 6.3 & 10.0 & 0.0145 & 0.0103 & 0.0076 \\
 & 336 & 4.3 & 6.0 & 9.1 & 0.0120 & 0.0098 & 0.0080 \\
 & 720 & 4.3 & 6.0 & 8.9 & 0.0106 & 0.0089 & 0.0069 \\
\midrule
ETTm1 & 96 & 3.4 & 4.0 & 6.4 & 0.0073 & 0.0061 & 0.0052 \\
 & 192 & 3.8 & 5.0 & 7.9 & 0.0095 & 0.0083 & 0.0064 \\
 & 336 & 4.1 & 5.6 & 9.2 & 0.0128 & 0.0096 & 0.0074 \\
 & 720 & 3.0 & 4.5 & 6.3 & 0.0075 & 0.0062 & 0.0055 \\
\midrule
ETTm2 & 96 & 4.1 & 5.8 & 9.4 & 0.0125 & 0.0095 & 0.0072 \\
 & 192 & 4.5 & 5.9 & 8.7 & 0.0128 & 0.0094 & 0.0072 \\
 & 336 & 4.1 & 5.8 & 8.5 & 0.0108 & 0.0082 & 0.0063 \\
 & 720 & 4.0 & 5.9 & 8.5 & 0.0138 & 0.0098 & 0.0074 \\
\bottomrule
\end{tabular}
}
\label{tab:sae_fidelity}
\end{table}

\section{SAE Sparsity Calibration}
\label{app:pareto}
 
To verify that our fixed sparsity penalty ($\lambda = 0.01$) does not artificially suppress valid features, we swept $\lambda \in \{0.1, 0.01, 0.001, 0.0001\}$ for both the $0.5\times$ and $4.0\times$ dictionaries across all eight benchmarks and four prediction horizons (256 SAE training runs total). Table~\ref{tab:pareto_summary} reports average $L_0$ (active latents per token) as a function of $\lambda$, averaged across horizons. As $\lambda$ decreases, $L_0$ increases smoothly across all datasets for both dictionary sizes, confirming that the SAE responds appropriately to reduced sparsity pressure. Notably, the $4.0\times$ dictionary consistently achieves higher $L_0$ than the $0.5\times$ dictionary at each $\lambda$, demonstrating that the larger dictionary does utilize some additional capacity --- yet as shown in our main results (Table~\ref{tab:sae_results}), this additional activation does not translate into downstream forecasting improvements. The combination of proper SAE calibration (this table) with flat downstream scaling (Table~\ref{tab:sae_results}) and dead latent patterns observed across the full test set (Table~\ref{tab:sae_results}) confirms that the absence of recoverable superposition is genuine, not an artifact of the sparsity penalty.
 
\begin{table}[H]
\centering
\caption{SAE sparsity calibration: average $L_0$ (active latents per token) across $\lambda$ values, averaged over all prediction horizons. Both dictionary sizes respond smoothly to reduced sparsity pressure, confirming the SAE instrument is properly calibrated.}
\footnotesize
\setlength{\tabcolsep}{4pt}
\begin{tabular}{l c c c c c c c c c}
\toprule
 & & \multicolumn{4}{c}{\textbf{$0.5\times$ Dictionary $L_0$}} & \multicolumn{4}{c}{\textbf{$4.0\times$ Dictionary $L_0$}} \\
\cmidrule(lr){3-6} \cmidrule(lr){7-10}
\textbf{Dataset} & \textbf{$d_\text{ff}$} & $\lambda{=}0.1$ & $\lambda{=}0.01$ & $\lambda{=}0.001$ & $\lambda{=}0.0001$ & $\lambda{=}0.1$ & $\lambda{=}0.01$ & $\lambda{=}0.001$ & $\lambda{=}0.0001$ \\
\midrule
Weather     & 128 & 0.0 & 3.7  & 19.8 & 53.5  & 0.0 & 5.3  & 32.6  & 143.5 \\
Electricity & 256 & 0.0 & 1.7  & 20.5 & 84.1  & 0.0 & 2.4  & 32.6  & 176.1 \\
Traffic     & 192 & 0.0 & 0.9  & 12.7 & 55.5  & 0.0 & 1.9  & 18.7  & 95.8  \\
Exchange    & 32  & 0.0 & 3.5  & 10.0 & 15.2  & 0.1 & 7.3  & 23.8  & 60.6  \\
ETTh1       & 32  & 0.0 & 3.8  & 12.0 & 15.2  & 0.0 & 8.0  & 28.3  & 69.5  \\
ETTh2       & 32  & 0.0 & 4.5  & 11.9 & 15.3  & 0.0 & 9.3  & 33.1  & 79.5  \\
ETTm1       & 32  & 0.0 & 3.6  & 11.4 & 15.3  & 0.0 & 7.4  & 26.2  & 65.1  \\
ETTm2       & 32  & 0.0 & 4.0  & 12.2 & 15.6  & 0.0 & 8.8  & 30.5  & 75.3  \\
\bottomrule
\end{tabular}
\label{tab:pareto_summary}
\end{table}
 
% ============================================================
% 5. DISCUSSION
% ============================================================
\section{Discussion}

Our goal is not to demonstrate that all time series transformers are free of superposition --- such a claim would require exhaustive analysis across every architecture and is neither feasible nor necessary. Instead, we demonstrate that superposition is \emph{not necessary} for competitive forecasting performance on standard benchmarks. PatchTST remains one of the strongest forecasting architectures, with recent large-scale evaluations confirming that no single model consistently dominates \citep{brigato2026champions}. If strong superposition were functionally important for these tasks, it should be detectable in the architecture that most closely mirrors the NLP transformers where superposition has been extensively documented.

The single-layer sufficiency result carries implications beyond architecture design. A task solvable by one transformer layer with $d_\text{model}$ as small as 16 has very low intrinsic dimensionality. The superposition hypothesis \citep{elhage2022superposition} predicts that superposition arises when a model must compress more features than it has dimensions. When there are fewer features than dimensions, no compression pressure exists. Our SAE results confirm this: expanding the dictionary does not uncover substantial hidden structure, overcomplete dictionaries provide no functional benefit, and improved reconstruction fidelity (Table~\ref{tab:sae_fidelity}) does not translate into downstream gains.

This stands in stark contrast to NLP, where depth serves a clear functional purpose. \citet{tenney2019bert} showed that BERT's layers resolve linguistic abstractions progressively --- from POS tagging to coreference --- and even \citet{niu2022does}, who challenged this pipeline interpretation, found rich linguistically motivated structure distributed across layers and token positions. Time series forecasting on standard benchmarks exhibits no analogous hierarchy: a single layer suffices, and its representations show no evidence of dense superimposed structure.

The persistent competitiveness of simple models --- DLinear \citep{zeng2023dlinear}, and more strikingly FITS \citep{xu2024fits}, which matches PatchTST using only ${\sim}10$k parameters via a single complex-valued frequency-domain layer --- follows naturally from this picture. \citet{chen2023tsmixer} reinforce this from the data side: their TMix-Only ablation shows that cross-variate mixing provides essentially no benefit on these benchmarks, suggesting the data lacks the cross-feature interactions that complex architectures are designed to exploit. If the underlying data does not require complex feature composition, transformer capacity goes largely unused, and progressively simpler models capture what there is to capture. The question may be better reframed: do standard forecasting benchmarks contain the representational richness that would justify transformer complexity? Our results suggest they do not, and the field may benefit from developing benchmarks with greater intrinsic complexity to meaningfully differentiate architectural advances.
% ============================================================
% 6. LIMITATIONS
% ============================================================
\section{Limitations}

We probe only the post-GELU intermediate FFN activations, a commonly used hook location in LLM SAE studies. Attention heads and residual stream representations may encode additional structure not captured by this analysis and represent a natural direction for future work. However, the FFN intermediate is the network's most expanded working space (2$\times$ the residual stream width), where sparse features would be expected to disentangle most easily and superposition pressure should be weakest. The absence of strong superposition signatures even in this expanded representation is therefore informative about the network's overall compression regime. Our goal is not to exhaustively characterize every transformer component, but to use mechanistic analysis to probe the representational demands of the underlying forecasting tasks themselves. The absence of strong FFN superposition suggests that these benchmarks do not exert the kind of feature-compression pressure commonly associated with transformer advantages in language modeling, helping explain the persistent competitiveness of simple linear models such as DLinear.

% ============================================================
% 7. CONCLUSION
% ============================================================

\section{Conclusion}
 
We applied sparse autoencoders to probe the internal representations of PatchTST, providing the first mechanistic interpretability analysis of a time series forecasting transformer. After establishing that a single-layer configuration achieves competitive performance across eight standard benchmarks, we showed that its post-GELU FFN activations exhibit no evidence of strong superposition: dictionary expansion yields negligible downstream improvement ($0.214\%$ average), overcomplete dictionaries remain largely inactive, and causal amplification of dominant latents produces minimal forecast perturbation. These findings demonstrate that superposition is not necessary for competitive time series forecasting on standard benchmarks, offering a mechanistic explanation for the persistent competitiveness of simple linear models and suggesting that the representational demands of these tasks are fundamentally lower than those of language.
 
\section{Broader Impact and Future Work}
 
While our analysis reveals that standard forecasting benchmarks do not demand complex sparse representations, this finding points toward domains where the opposite may hold. Medical time series --- such as EEG recordings preceding epileptic seizures, ECG signals before cardiac events, or intracranial pressure traces prior to stroke --- may exhibit precisely the kind of representational complexity that these benchmarks lack. Transformers trained on such data may need to compress subtle, clinically meaningful physiological patterns into their representations, potentially giving rise to genuine superposition.
 
In such settings, the SAE methodology demonstrated here could serve as a bridge between model predictions and clinical understanding. If a transformer reliably forecasts an impending seizure, SAE decomposition could isolate the specific latent features that activate in the pre-ictal period. These latents could then be mapped back to input channels and temporal windows, producing candidate biomarkers for neurologists to investigate. Rather than treating the model as an opaque predictor, clinicians could use SAE-derived features as interpretable hypotheses about physiological correlates that are invisible to unaided observation.

Financial time series present another promising domain. Asset returns exhibit regime-dependent dynamics, cross-instrument contagion, and macroeconomic dependencies that jointly create representational demands far exceeding those of the benchmarks studied here. Models such as Temporal Fusion Transformers \citep{lim2021temporal} have shown that interpretable attention improves both performance and practitioner trust in financial settings. SAE decomposition could reveal whether transformers trained on multi-asset portfolios compress regime indicators and cross-asset dependencies into superimposed directions, surfacing latent features that analysts could map to actionable risk factors.

\section*{Code Availability}

The complete codebase for reproducing the experiments in this study is publicly available at \url{https://github.com/AlperYildirim1/TS-MechInterp}. 
% =====
\newpage
% ============================================================
% REFERENCES
% ============================================================
\bibliographystyle{plainnat}
\bibliography{references}

\appendix

\section{Single-Layer Base Model Performance}
\label{app:single_layer}

Table~\ref{tab:single_layer} compares our single-layer model against the published PatchTST/42 results \citep{nie2023patchtst}, which use 3 transformer layers together with extensive per-dataset tuning. On Weather and Electricity, the results are nearly identical, while on ETTm1 and ETTm2 the differences remain modest. Larger gaps appear on Traffic and ETTh2, likely reflecting the absence of dataset-specific tuning rather than a clear failure of shallow architectures. Notably, Traffic exhibits over $60\%$ dead latents even at the larger $d_\text{model}=96$ configuration used in our experiments, suggesting that increasing representational capacity alone does not substantially benefit this dataset.

Taken together, these results indicate that strong forecasting performance on standard benchmarks does not appear to require substantial transformer depth or highly specialized configurations.

\begin{table}[H]
\centering
\caption{Forecasting performance comparison between our single-layer model (unified configuration, no per-dataset tuning) and the published PatchTST/42 \citep{nie2023patchtst} (3 layers, per-dataset tuning). Exchange is excluded as it was not evaluated in the original PatchTST paper.}
\footnotesize
\setlength{\tabcolsep}{4pt}
\begin{tabular}{l c c c c c}
\toprule
 & & \multicolumn{2}{c}{\textbf{PatchTST/42} \citep{nie2023patchtst}} & \multicolumn{2}{c}{\textbf{Ours (1 Layer)}} \\
\cmidrule(lr){3-4} \cmidrule(lr){5-6}
\textbf{Dataset} & \textbf{Horizon} & \textbf{MSE} & \textbf{MAE} & \textbf{MSE} & \textbf{MAE} \\
\midrule
Weather & 96  & 0.152 & 0.199 & 0.148 & 0.199 \\
        & 192 & 0.197 & 0.243 & 0.193 & 0.244 \\
        & 336 & 0.249 & 0.283 & 0.248 & 0.285 \\
        & 720 & 0.320 & 0.335 & 0.322 & 0.338 \\
\midrule
Electricity & 96  & 0.130 & 0.222 & 0.130 & 0.226 \\
            & 192 & 0.148 & 0.240 & 0.149 & 0.242 \\
            & 336 & 0.167 & 0.261 & 0.168 & 0.262 \\
            & 720 & 0.202 & 0.291 & 0.205 & 0.296 \\
\midrule
Traffic & 96  & 0.367 & 0.251 & 0.390 & 0.263 \\
        & 192 & 0.385 & 0.259 & 0.417 & 0.293 \\
        & 336 & 0.398 & 0.265 & 0.428 & 0.297 \\
        & 720 & 0.434 & 0.287 & 0.457 & 0.310 \\
\midrule
ETTh1 & 96  & 0.375 & 0.399 & 0.378 & 0.398 \\
      & 192 & 0.414 & 0.421 & 0.423 & 0.424 \\
      & 336 & 0.431 & 0.436 & 0.452 & 0.446 \\
      & 720 & 0.449 & 0.466 & 0.477 & 0.486 \\
\midrule
ETTh2 & 96  & 0.274 & 0.336 & 0.292 & 0.352 \\
      & 192 & 0.339 & 0.379 & 0.356 & 0.392 \\
      & 336 & 0.331 & 0.380 & 0.384 & 0.413 \\
      & 720 & 0.379 & 0.422 & 0.409 & 0.440 \\
\midrule
ETTm1 & 96  & 0.290 & 0.342 & 0.307 & 0.350 \\
      & 192 & 0.332 & 0.369 & 0.338 & 0.368 \\
      & 336 & 0.366 & 0.392 & 0.366 & 0.393 \\
      & 720 & 0.420 & 0.424 & 0.429 & 0.418 \\
\midrule
ETTm2 & 96  & 0.165 & 0.255 & 0.171 & 0.260 \\
      & 192 & 0.220 & 0.292 & 0.227 & 0.298 \\
      & 336 & 0.278 & 0.329 & 0.289 & 0.337 \\
      & 720 & 0.367 & 0.385 & 0.371 & 0.395 \\
\bottomrule
\end{tabular}
\label{tab:single_layer}
\end{table}

A natural question is whether the modest performance gaps on datasets such as Traffic and ETTh2 reflect representational structure that our single-layer model fails to capture. While we cannot rule out that closing the final ${\sim}2\%$ MSE gap might involve richer internal representations, including potentially superposition, the burden of this argument is considerable: it would require that superposition contributes nothing to the first $98\%$ of performance and emerges only in the narrow margin between our model and the tuned baseline. We consider this unlikely, but emphasize that our claim does not depend on resolving this question. The demonstrated result --- that competitive forecasting performance is achievable without detectable superposition --- stands regardless.

% ============================================================
% APPENDIX B: FFN Functional Validation
% Add after existing Appendix A (Single-Layer Base Model Performance)
% ============================================================

\section{FFN Functional Validation}
\label{app:ffn_ablation}

A prerequisite for interpreting SAE-based superposition analysis is confirming that the probed component performs meaningful computation. If the post-GELU FFN activations were functionally inert, the absence of superposition signatures would be trivially explained --- one cannot find compressed features in a component that contributes nothing to the output. To rule this out, we perform a zero ablation: during inference, all post-GELU intermediate activations are replaced with zeros, effectively removing the FFN's additive contribution to the residual stream while leaving the attention output and skip connection intact.

Table~\ref{tab:zero_ablation} reports the resulting forecasting degradation across all eight benchmarks and four prediction horizons. The FFN is clearly functionally active, with zero ablation producing substantial MSE increases in the majority of settings. The effect is most pronounced on the higher-dimensional datasets: Electricity exhibits degradation of +131.4\% at $H$=96 and Traffic +40.3\%, reflecting the greater computational load carried by their wider FFN layers ($d_\text{ff}$ = 256 and 192 respectively). Weather shows a similar pattern, with +55.2\% degradation at the shortest horizon. Even the narrow-dimensional ETT benchmarks ($d_\text{ff}$ = 32) show consistent positive degradation in most configurations, typically ranging from +1\% to +8\%.

A small number of dataset-horizon pairs exhibit near-zero or slightly negative degradation, most notably Exchange at $H$=192 ($-6.6\%$) and ETTh2 at $H$=336 and 720 ($-0.9\%$, $-0.7\%$). These cases suggest that for certain configurations, the FFN's contribution is marginal or mildly harmful --- consistent with our broader finding that these benchmarks do not demand substantial nonlinear feature composition. Importantly, even in these cases, the SAE analysis in Section~4 remains informative: the dictionary scaling and causal intervention results characterize whatever structure the FFN \textit{does} encode, whether or not that structure is net-beneficial at every horizon.

Across the full set of 32 evaluations, 28 show positive degradation upon FFN removal, with a median degradation of +6.8\%. This confirms that the post-GELU hook is a suitable and informative site for SAE-based analysis: the representations being probed are, in the large majority of settings, functionally active, and the superposition findings reported in Section~4 reflect properties of a working component.

\begin{table}[h]
\centering
\caption{FFN zero ablation: forecasting degradation when post-GELU activations are replaced with zeros during inference. The FFN is functionally active across the large majority of settings, with degradation most pronounced on higher-dimensional datasets.}
\label{tab:zero_ablation}
\small
\begin{tabular}{llccc}
\toprule
Dataset & Horizon & Base MSE & Ablated MSE & Degradation (\%) \\
\midrule
Weather       & 96  & 0.1480 & 0.2297 & $+55.2$ \\
              & 192 & 0.1927 & 0.2365 & $+22.7$ \\
              & 336 & 0.2482 & 0.2656 & $+7.0$ \\
              & 720 & 0.3221 & 0.3599 & $+11.7$ \\
\midrule
Electricity   & 96  & 0.1303 & 0.3015 & $+131.4$ \\
              & 192 & 0.1491 & 0.2708 & $+81.7$ \\
              & 336 & 0.1680 & 0.2540 & $+51.2$ \\
              & 720 & 0.2049 & 0.2569 & $+25.4$ \\
\midrule
Traffic       & 96  & 0.3895 & 0.5465 & $+40.3$ \\
              & 192 & 0.4174 & 0.4754 & $+13.9$ \\
              & 336 & 0.4282 & 0.4806 & $+12.3$ \\
              & 720 & 0.4566 & 0.4793 & $+5.0$ \\
\midrule
Exchange      & 96  & 0.0963 & 0.1099 & $+14.1$ \\
              & 192 & 0.1911 & 0.1785 & $-6.6$ \\
              & 336 & 0.3723 & 0.3801 & $+2.1$ \\
              & 720 & 0.9783 & 1.0428 & $+6.6$ \\
\midrule
ETTh1         & 96  & 0.3779 & 0.4045 & $+7.0$ \\
              & 192 & 0.4230 & 0.4332 & $+2.4$ \\
              & 336 & 0.4524 & 0.4636 & $+2.5$ \\
              & 720 & 0.4774 & 0.4841 & $+1.4$ \\
\midrule
ETTh2         & 96  & 0.2923 & 0.3132 & $+7.2$ \\
              & 192 & 0.3562 & 0.3587 & $+0.7$ \\
              & 336 & 0.3839 & 0.3803 & $-0.9$ \\
              & 720 & 0.4090 & 0.4063 & $-0.7$ \\
\midrule
ETTm1         & 96  & 0.3070 & 0.3108 & $+1.2$ \\
              & 192 & 0.3384 & 0.3503 & $+3.5$ \\
              & 336 & 0.3655 & 0.3717 & $+1.7$ \\
              & 720 & 0.4288 & 0.4347 & $+1.4$ \\
\midrule
ETTm2         & 96  & 0.1710 & 0.1770 & $+3.5$ \\
              & 192 & 0.2269 & 0.2458 & $+8.3$ \\
              & 336 & 0.2885 & 0.2938 & $+1.8$ \\
              & 720 & 0.3706 & 0.3712 & $+0.2$ \\
\bottomrule
\end{tabular}
\end{table}

\end{document}